\documentclass[5p]{elsarticle}
\usepackage{color}
\usepackage{amsmath,amsfonts,amssymb}
\usepackage{multirow}
\usepackage{subcaption}

\usepackage{hyperref}

\journal{Information Fusion}

\begin{document}

\begin{frontmatter}

\title{Multi-level Attention Fusion Network for Audio-visual Event Recognition}


\author[sherbrooke,mons]{Mathilde Brousmiche \corref{mycorrespondingauthor}}
\cortext[mycorrespondingauthor]{Corresponding author}
\ead{mathilde.brousmiche@umons.ac.be, mathilde.brousmiche@usherbrooke.ca}

\author[sherbrooke]{Jean Rouat}

\author[mons]{Stéphane Dupont}

\address[sherbrooke]{University of Sherbrooke, NECOTIS Lab,  2500 boul. de l'Université, Sherbrooke, Canada}
\address[mons]{University of Mons, Numediart Institute, 20 Place du Parc, Mons, Belgium}




\begin{abstract}

Event classification is inherently sequential and multimodal. Therefore, deep neural models need to dynamically focus on the most relevant time window and/or modality of a video. In this study, we propose the Multi-level Attention Fusion network (MAFnet), an architecture that can dynamically fuse visual and audio information for event recognition. Inspired by prior studies in neuroscience, we couple both modalities at different levels of visual and audio paths. Furthermore, the network dynamically highlights a modality at a given time window relevant to classify events. Experimental results in AVE (Audio-Visual Event), UCF51, and Kinetics-Sounds datasets show that the approach can effectively improve the accuracy in audio-visual event classification. Code is available at: \url{https://github.com/numediart/MAFnet}
\end{abstract}


\end{frontmatter}


\section{Introduction}

Event recognition is an active research area in machine learning. It has numerous potential applications such as video surveillance \cite{dufour2012intelligent}, autonomous driving \cite{kala2016road}, sports analysis \cite{d2010review}, and content-based retrieval \cite{gibbon2008introduction}. Thanks to the collection of large video datasets such as YouTube-8M \cite{real2017youtube}, Kinetics \cite{carreira2017quo} and Sports-1M \cite{karpathy2014large}, event recognition performance has improved during recent years. However, compared to the success of image classification, event recognition is still a challenging task due to the high computational complexity. The current state-of-the-art performance remains not as accurate as human performance.

Inspired by the success of image classification \cite{simonyan2014very, szegedy2016rethinking}, current models for visual event recognition use convolutional neural networks (CNN). Several methods have been explored to process the temporal information such as two-streams network \cite{simonyan2014two}, recurrent neural network (RNN) \cite{donahue2015long}, and three-dimensional convolution  (3D CNN) \cite{tran2015learning}, to name a few. However, all these works ignore an important information present in video: acoustic features.

Video understanding is a natural human ability. From an early age, humans are able to understand events or actions based on image, video as well as sound. However, most contemporary approaches ignore the acoustic information. It is obvious that the acoustic signal can be useful for event recognition. For example, two musical instruments may be difficult to distinguish based on the visual information but can produce distinct sounds or the object of interest may be occluded. Furthermore, some actions such as \textit{Whistling} are visually subtle but can be recognized based on acoustic features. 

Given the potential of sound to facilitate event recognition, researchers have attempted to combine the audio and visual signals \cite{kazakos2019epic, arandjelovic2017look, arandjelovic2018objects, aytar2016soundnet, owens2018audio, owens2016ambient, tian2018audio, tian2019multi, long2018multimodal}. Still, audio-visual event recognition is not easy as it is difficult to effectively use the audio information. In fact, audio can be corrupted with irrelevant background noise and sounds. Moreover, some event such as \textit{Shaking hands} does not produce a particular sound signature. 

Visual and audio information are different types of signals. Therefore, how to exploit the maximum relevant information coming from both modalities? There is not a simple answer. In real life situations, scenes are dynamics and the respective contribution of audio and video to the scene (or object) understanding evolves through time. Effective solutions should be adaptive and take into consideration the fact that the dynamics of visual and acoustics inputs are very different and change with the actions in the scene.

In the context of deep learning, visual and audio paths do not have the same complexity and therefore the same learning dynamics during training. Indeed, the visual and audio paths do not have the same learning speed, the number of iterations necessary to train a network based on visual information is not the same as for a network based on audio. This can lead to a generalization problem when training modalities together.

Therefore, we propose the Multi-level Attention Fusion network (MAFnet) to fuse dynamically visual and audio information for event recognition. This network dynamically associates a score to each modality and time window in videos. The score highlights a modality at a given time window that may be effective to recognize the event. We also propose to go further than the simple fusion by coupling modalities with a lateral connection between visual and audio paths of MAFnet. Moreover, to overcome the incompatibility of learning dynamics between visual and audio paths, we propose to randomly drop the update of the visual path during training. We evaluate our architecture on multiple datasets: Kinetics-Sound \cite{carreira2017quo}, UCF51 \cite{soomro2012ucf101} and AVE \cite{tian2018audio}. In addition, we evaluate the contribution of each module of MAFnet with an ablation study.


\section{Related work}
\subsection{Visual event recognition}
Inspired by the success of image classification, convolutional neural networks (CNN) have also been applied in visual event recognition. Several methods have been proposed to take advantage of the temporal information. The usage of recurrent neural networks (RNN) on top of 2D convolutional layers is investigated in \cite{donahue2015long, yue2015beyond, ma2019ts} to take into account long-term dependencies. Li \textit{et al.} went further by proposing a convolutional long-short term memory (LSTM) \cite{li2018videolstm}. Another approach was to extend 2D convolution kernels to 3D convolution kernels to learn spatio-temporal features \cite{karpathy2014large, tran2015learning, carreira2017quo}. In addition, to reduce complexity, the 3D convolution is decomposed into two convolutions: a spatial 2D convolution and a temporal 1D convolution \cite{tran2018closer, xie2018rethinking}. Another strategy was to capture fine low-level motion by calculating optical flow \cite{simonyan2014two, feichtenhofer2016convolutional, wang2016temporal}. However, all these techniques do not exploit an important part of the video classification: the acoustic information.

\subsection{Audio-visual event recognition}

In recent years, only few works exploited the information present in the audio signal. The concatenation is used in \cite{long2018attention, kazakos2019epic} to fuse the visual and audio paths to exploit the information from the two modalities. Long \textit{et al.} went further by testing different levels of fusion in the network \cite{long2018multimodal}. These networks integrated the visual and audio information with hard fusion without exploiting a possible interaction between visual and audio paths. These works did not study more complex fusion techniques such as multimodal compact bilinear pooling (MCB) \cite{gao2016compact} or dual multimodal residual fusion (DMR) \cite{tian2018audio}. Furthermore, they did not take into account the different learning dynamics of the different modalities.

\subsection{Audio-visual event detection}
The release of the AVE dataset \cite{tian2018audio} has stimulated research in audio-visual event detection. For example,  an audio-guided visual attention mechanism is introduced in \cite{tian2018audio} to learn which visual region to look at based on the visual and audio information. Lin \textit{et al.} proposed to learn global and local event information in a sequence to sequence manner \cite{lin2019dual}. Finally, Wu \textit{et al.} extracted the global representation of one modality and found the local segments that are relevant to the event in the other modality and vice versa \cite{wu2019dual}. In our work, we propose to better integrate audio-visual information by computing a global feature with an attention module to include only the relevant information present in the modalities. In addition, instead of visual spatial attention, we propose to use an attention mechanism on the audio feature maps based on the visual information, called modality conditioning. Indeed, we observed that the visual modality has more information to contribute to the audio path of MAFnet than the reverse.

\subsection{Modality conditioning}

Modality conditioning is the influence of a modality on another modality. It is the interaction between the paths of each modality inside the neural network. Interactions can be created by simple operations between paths such as an element-wise multiplication \cite{tian2019multi} or a sum \cite{xiao2020audiovisual} at different levels of the network.

More complex approaches to condition modalities have been explored, for example, the attention mechanism. Attention models were first proposed for object detection \cite{mnih2014recurrent} and then used for other applications such as natural language processing with the self-attention mechanism \cite{vaswani2017attention}. Attention has been applied to video classification under the form of temporal and/or spatial attention \cite{wang2020cascade, li2020spatio, long2018multimodal}. However, these models did not include an interaction between modality paths. Only Tian \textit{et al.} proposes to realize the interaction between paths by implementing visual attention guided by audio to condition vision \cite{tian2018audio}.

Another approach to condition one modality with the other is the conditional normalization (CN). Instead of focusing attention on a particular region of space or a particular time window, the CN highlights some feature maps based on a given input. Various forms of CN have proven to be highly effective across a number of domains and modalities: image stylization \cite{huang2017arbitrary}, speech recognition \cite{kim2017dynamic}, visual question answering \cite{de2017modulating} and audio question answering \cite{abdelnour2019visual}. As presented in our previous work \cite{brousmiche2019audio}, we propose to condition the audio path with the visual information by using the Feature-wise Linear Modulation (FiLM) method \cite{perez2018film}. The FiLM layer highlights audio feature maps based on visual information.


\section{Multi-level Attention Fusion network}

\begin{figure*}[ht]
    \centering
    \includegraphics[width=\textwidth]{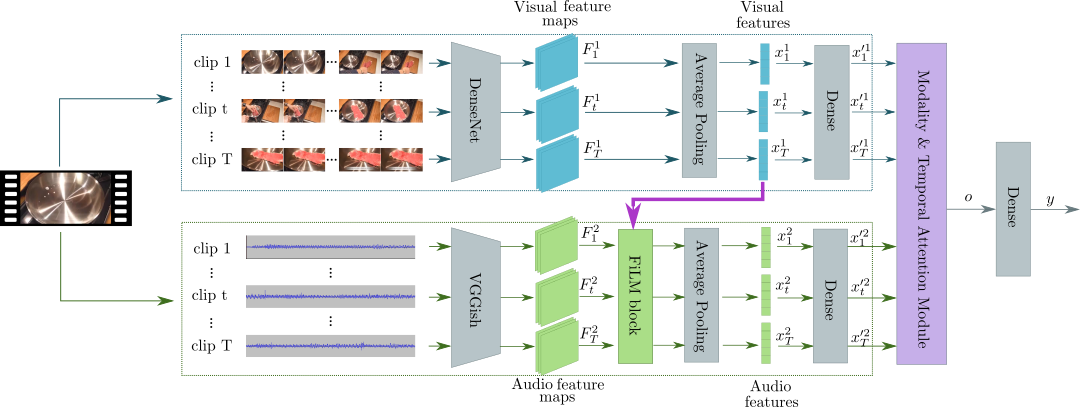}
    \caption{Multi-level Attention Fusion network (MAFnet): one video is split into T non-overlapping clips. Then, audio and visual information are extracted with two pretrained CNNs: DenseNet \cite{huang2017densely} for visual features and VGGish \cite{hershey2017cnn} for audio features. The clip features are further fed into the modality \& temporal attention module to build a global feature comprising multimodal and temporal information. This global feature is then used to predict the label of the video. A lateral connection between visual and audio paths is created trough the FiLM layer \cite{perez2018film}. }
    \label{fig:model}
\end{figure*}

Inspired by the ability of humans to pay attention to different regions, instants and modalities \cite{goldstein2016sensation}, we propose to compute a score for each modality and for each time window with the modality \& temporal attention module. The attention module combines modality and temporal information to create a global feature containing the relevant multimodal and temporal information.

In addition to the fusion with the attention module, we propose to go further than modality fusion at high level and include interaction between visual and audio paths with a FiLM layer. In this section, we overview the Multi-level Attention Fusion network (MAFnet) and then detail the different components of the network.

\subsection{Overview of the Multi-level Attention Fusion network (MAFnet)}


Fig. \ref{fig:model} presents the architecture of MAFnet. As in \cite{tian2018audio}, we split each video into $T$ non-overlapping clips, where each clip is 1s long. We extract information for $K=2$ modalities (visual and audio information). For each clip, we extract visual and audio feature maps with pretrained convolutional neural networks. So, we have 2 input sequences: $\{F^1_1,\dots, F^1_T \}, F_t^1 \in \mathbb{R}^{H_v \times W_v \times D_v}$ for the visual information and $\{F_1^2,\dots, F_T^2 \}, F_t^2 \in \mathbb{R}^{H_a \times W_a \times D_a}$ for the audio information. $H$, $W$ and $D$ are respectively the height, the width and the number of feature maps.

We reduce the feature maps with average pooling and feed the visual features (($\{x^1_1,\dots, x^1_T \}, x_t^1 \in \mathbb{R}^{D_v}$) and audio features ($ \{x^2_1,\dots, x^2_T \}, x_t^2 \in \mathbb{R}^{D_a} $) in the modality \& temporal attention module. This module is the combination of temporal and modality attentions. It attempts to learn the attention scores $\lambda_t^k$ with $t=1,\dots,T$ and $k=1,\dots,K$ to weight temporal and modality dimensions. We therefore obtain a temporal-multimodal representation of the entire video. The output of the network is $y \in \mathbb{R}^N$ with $N$ the number of classes.

To go further than a simple fusion, we implement a lateral connection between visual and audio paths with the FiLM layer \cite{perez2018film}. With the FiLM layer, the visual modality influences the audio modality. Greater importance is given to some of the audio feature maps based on the visual information. The FiLM layer is placed directly at the output of the audio feature extractor before reducing feature maps into vectors.


\begin{figure}[t]
\centering
\begin{subfigure}{.13\textwidth}
  \centering
  \includegraphics[width=1\textwidth]{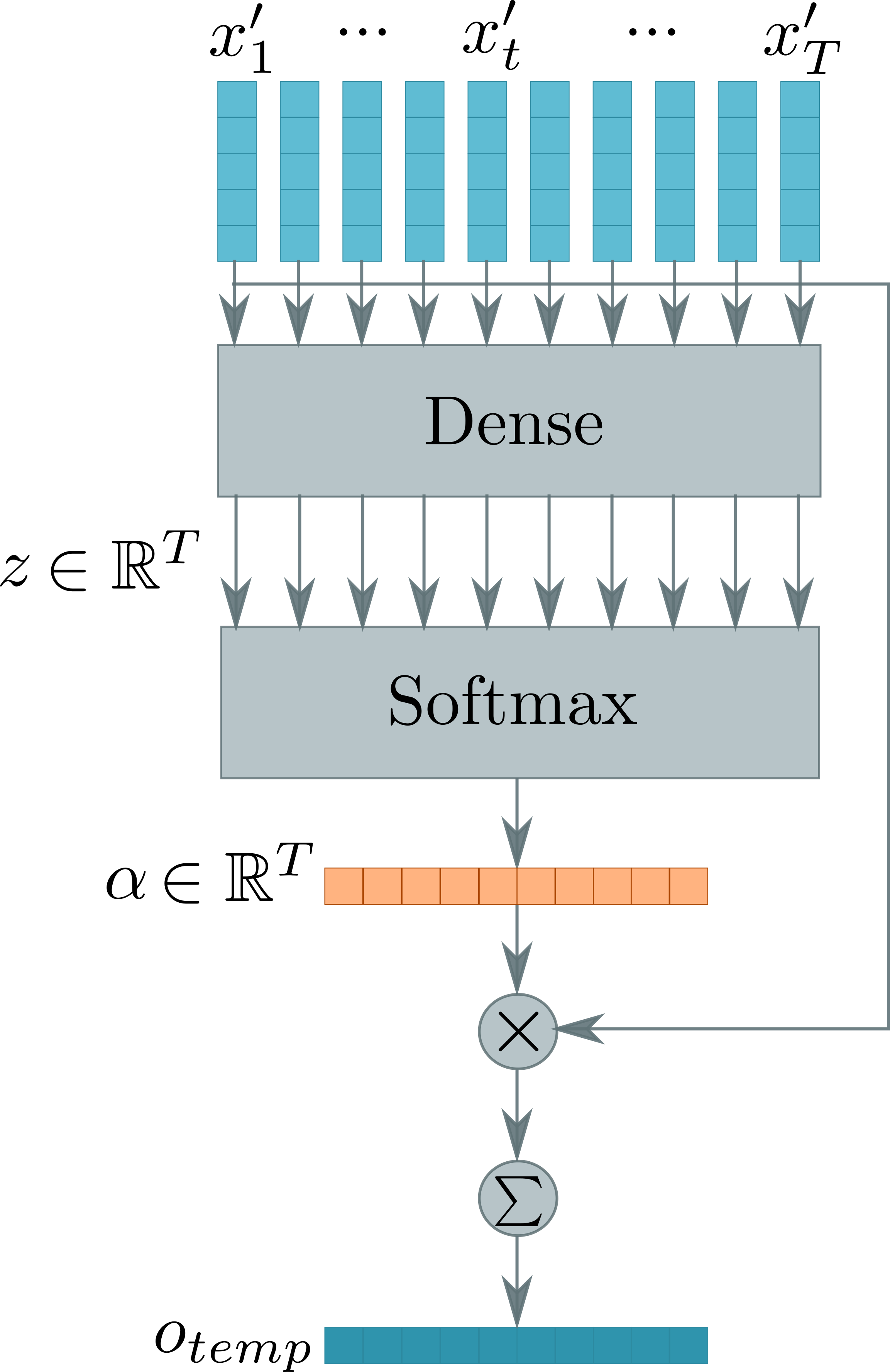}
  \caption{Temporal attention}
  \label{fig:sub1}
\end{subfigure}%
\quad
\begin{subfigure}{.13\textwidth}
  \centering
  \includegraphics[width=1\textwidth]{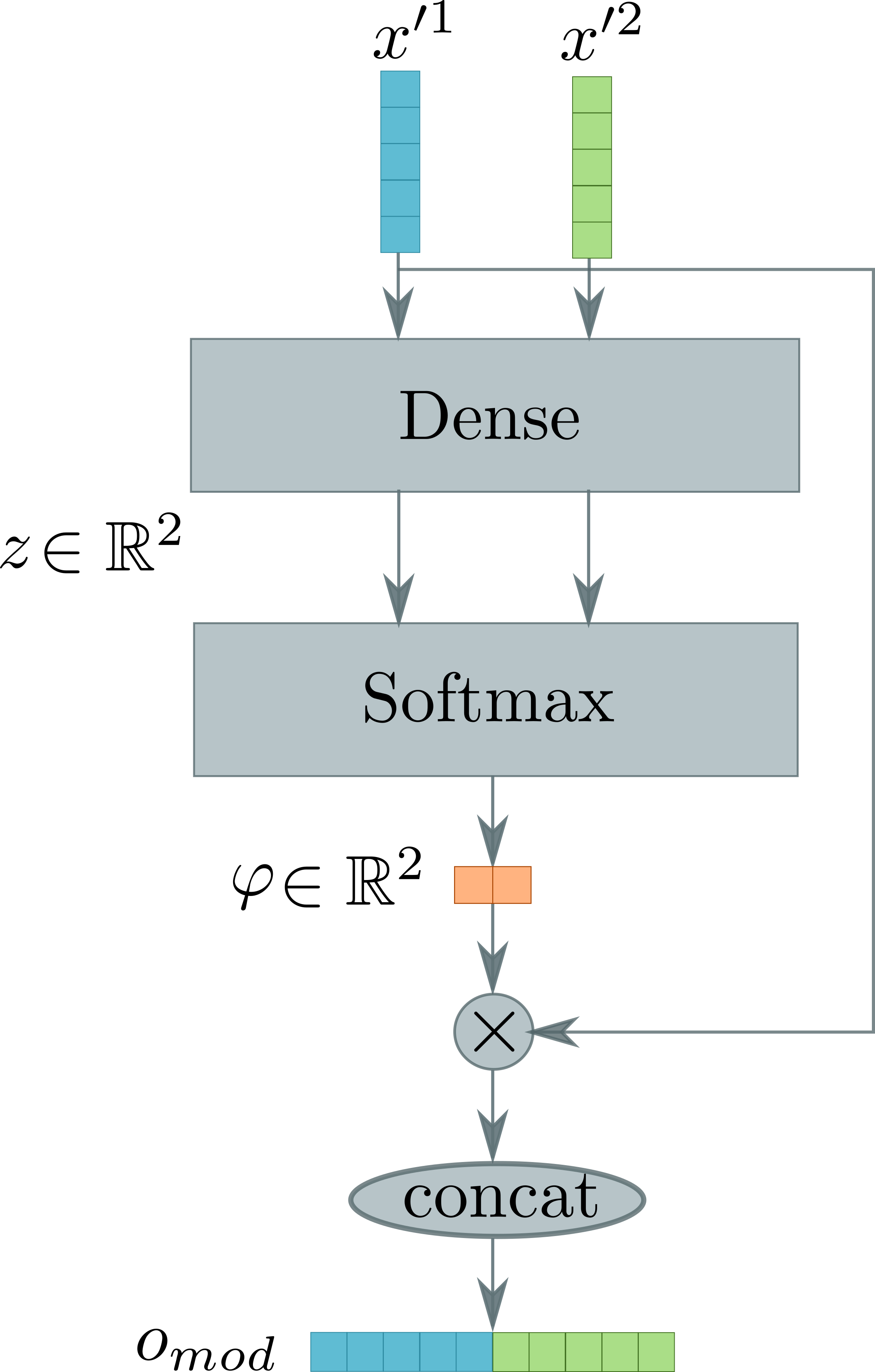}
  \caption{Modality attention}
  \label{fig:sub2}
\end{subfigure}%
\quad
\begin{subfigure}{.17\textwidth}
  \centering
  \includegraphics[width=1\textwidth]{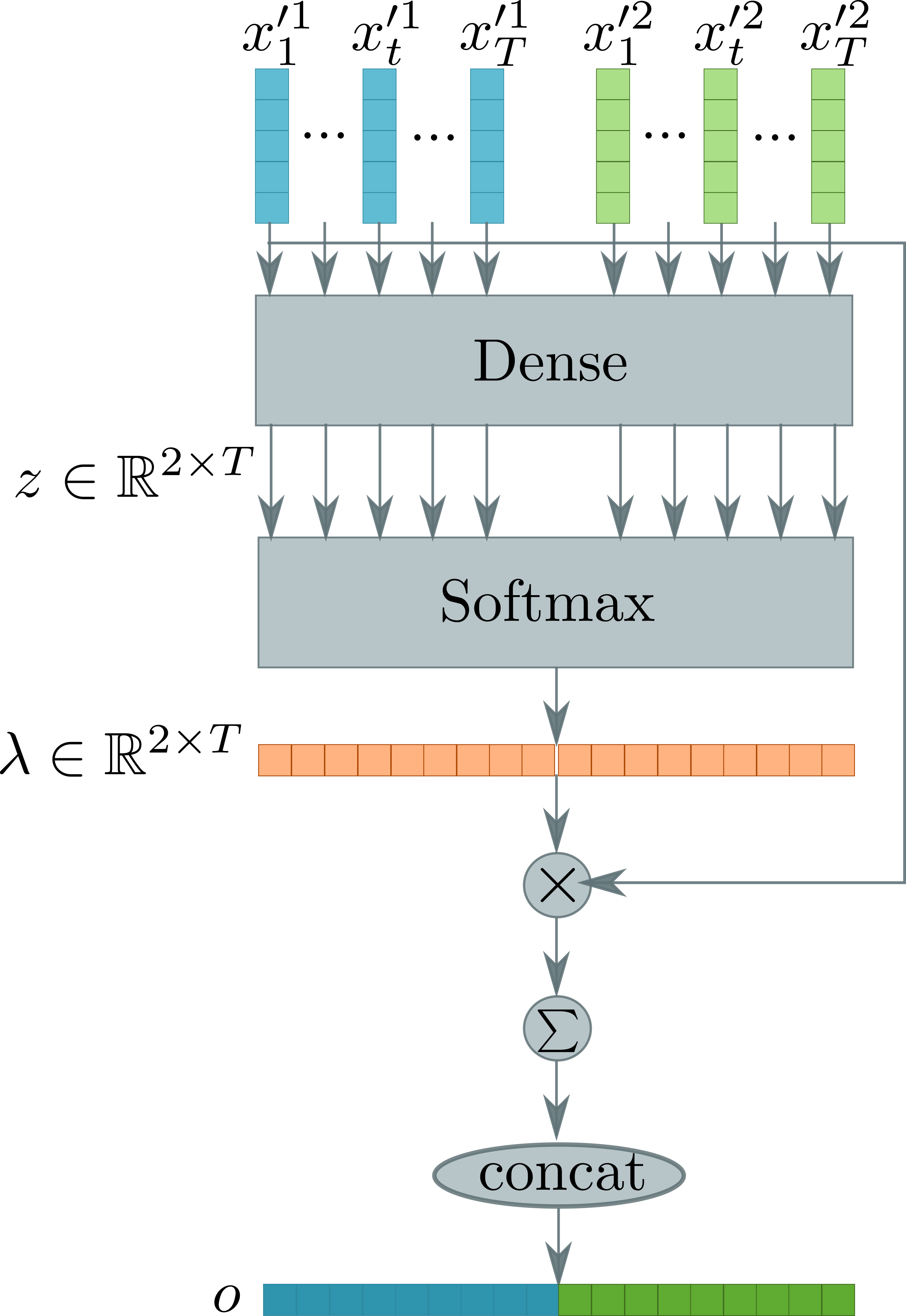}
  \caption{Temporal \& modality attention}
  \label{fig:sub3}
\end{subfigure}
\caption{Attention mechanisms. (a) Temporal attention: a score $\alpha$ is computed for each time window and the video-level feature representation $o_{temp}$ with the sum. (b) Modality attention: a score $\varphi$ is computed for each modality and the multimodal feature representation $o_{mod}$ with the concatenation. (c) Temporal \& modality attention: a score $\lambda$ is computed for each time window AND modality and the global feature representation $o$ with the combination of the sum over time windows and the concatenation over modalities.}
\label{fig:test}
\end{figure}

\subsection{Temporal attention}

The aim of temporal attention \cite{long2018multimodal, wu2019dual} is to assign a positive weight score to each clip descriptors extracted from the video (Fig. \ref{fig:sub1}). The score can be interpreted as the relative contribution of each clip to the recognition of the target action, or the relative importance of each clip to generate an accurate global video representation.

Technically, for a given modality, given the input feature $X' = \{x'_1,\dots, x'_T \}, x'_t \in \mathbb{R}^{D}$, the corresponding score $\alpha = \{\alpha_1, \dots, \alpha_T\}$ over the T feature vectors is computed by

\begin{equation}
    z_t = g_{att}(x'_t;\theta_{att}) = ReLU(W_{temp}^T x'_t + b)
\end{equation}

\begin{equation}
    \alpha_t = \frac{\exp(z_t)}{\sum_{j=1}^{T} \exp(z_j)}
\end{equation}
where $g_{att}$ is the temporal attention network parameterized by $\theta_{att}$. $g_{att}$ can take different forms such as a perceptron. $z_t$ is an intermediate attention score, normalized with the softmax function.

We compute the video-level feature representation $o_{temp}$ with the weighted sum of the clip features. The weights are the scores computed by the temporal attention module:
\begin{equation}
    o_{temp} = \sum_{t=1}^{T} \alpha_t x'_t
\end{equation}

\subsection{Modality attention}
In the context of speech recognition, Zhou \textit{et al.} proposed a modality attention mechanism \cite{zhou2019modality}. The attention mechanism fuses input from multiple modalities into a single representation by weighted summing the information from individual modalities. We propose to use a similar mechanism but use the concatenation of the weighted modalities instead of the sum (Fig. \ref{fig:sub2}). In subsection \ref{ssec:analysis}, we discuss the choice of the modality fusion. 

The attention module computes a score for each modality, the score is proportional to the importance of the modality for the video classification.

Technically, at a given time, given the input feature $X' = \{x'^1,\dots, x'^K \}, x^k \in \mathbb{R}^{D_k}$ with $K$ the number of modalities. The score for each modality is computed by:

\begin{equation}
    z^k = h_{att}(x'^k;\theta_{att}) = ReLU(W_{mod}^T x'^k + b)
\end{equation}

\begin{equation}
    \varphi^k = \frac{\exp(z^k)}{\sum_{j=1}^{K} \exp(z^j)}
\end{equation}
where $h_{att}$ is the attention network parameterized by $\theta_{att}$ and $z^k$ is an intermediate attention score, normalized with the softmax function.

The multimodal feature $o_{mod}$ is obtained by fusing the weighted unimodal features with a concatenation:

\begin{equation}
    o_{mod} = concat( [ \varphi^1 x'^1, \dots,  \varphi^K x'^K])
\end{equation}

The modality attention module can dynamically choose the most relevant modality for a better classification of the events. Indeed, we can imagine that \textit{Frying (food)} or \textit{Truck} have strong visual information while \textit{Violin} or \textit{Flute} have strong audio information. 

\begin{figure}[t]
    \centering
    \includegraphics[width=0.45\textwidth]{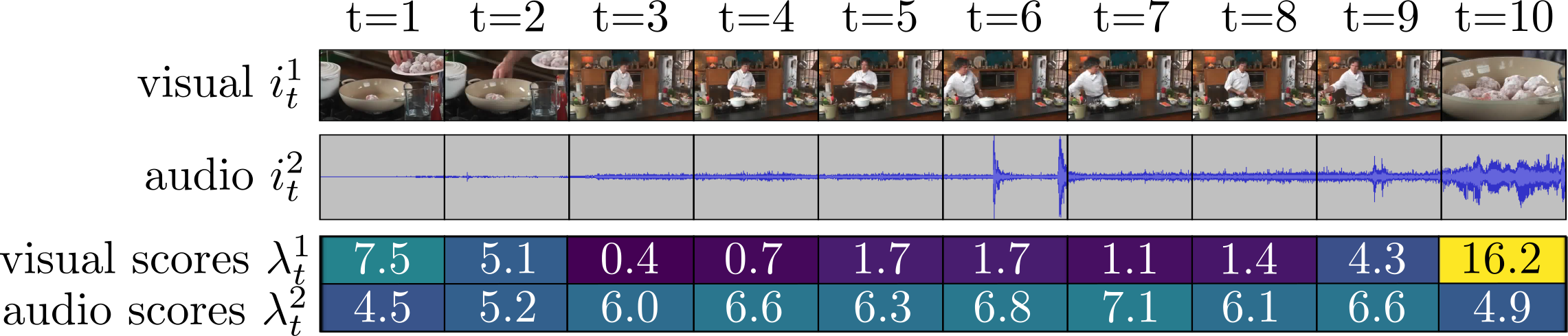}
    \caption{Visualization of the scores $\lambda_t^k$ determined by the modality \& temporal attention module for a video labeled \textit{Frying (food)} of the AVE dataset.}
    \label{fig:mod_temp_attention}
\end{figure}

\subsection{Modality \& temporal attention module}

We can combine the temporal and the modality attention modules to constitute the modality \& temporal attention module (Fig. \ref{fig:sub3}). The aim of the modality \& temporal attention module is to assign a positive score for each modality and clip. Indeed, for example in Fig. \ref{fig:mod_temp_attention}, we see that most of the time the audio information is more relevant than the visual information except for the last clip where you can clearly see the food frying. This visual clip has the largest score and can be identified as the most relevant to classify the video as \textit{Frying (food)}.

If we have the input:

\begin{equation}
    X' = \begin{bmatrix}
        x'^1_1 & \cdots & x'^1_t & \cdots & x'^1_T\\
          & \cdots &   & \cdots &  \\
        x'^K_1 & \cdots & x'^K_t & \cdots & x'^K_T
    \end{bmatrix}
\end{equation}

The equations of the attention module become:

\begin{equation}
    z_t^k = f_{att}(x'^k_t;\theta_{att})
          = ReLU(W_{mod+temp}^T x'^k_t + b) 
\end{equation}

\begin{equation}
    \lambda_t^k = \frac{\exp(z_t^k)}{\sum_{j=1}^{T} \sum_{l=1}^{K} \exp(z_j^l)}
\end{equation}

\begin{equation}
    o = concat( [ \sum_{t=1}^{T} \lambda_t^1 x'^1_t, \dots,  \sum_{t=1}^{T} \lambda_t^K x'^K_t])
\end{equation}

We add a dense layer in the path of each modality before the attention module because the attention module needs each modality to have the same dimension.

\subsection{Lateral connection}
\begin{figure}[t]
    \centering
    \includegraphics[width=0.4\textwidth]{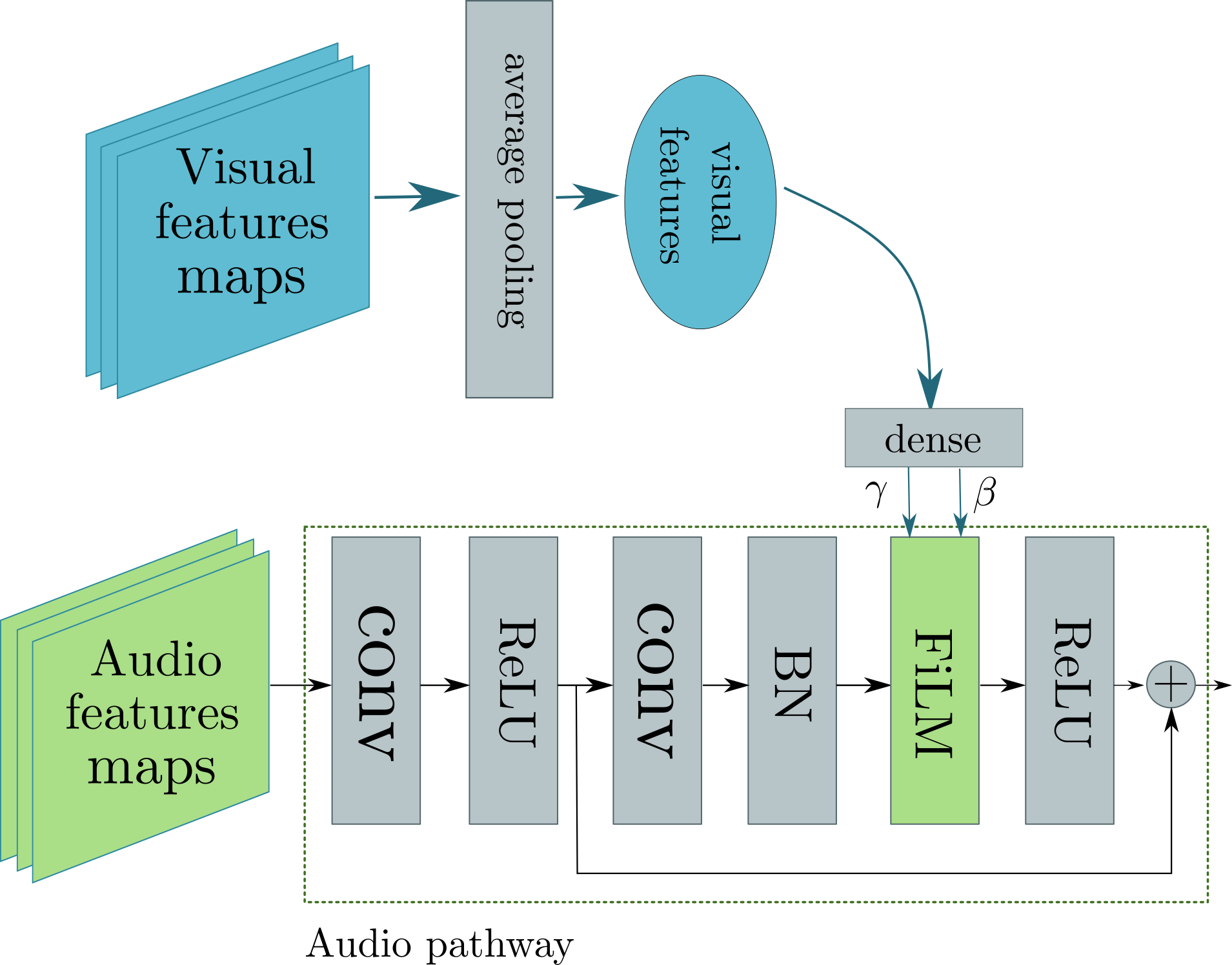}
    \caption{Lateral connection between visual and audio paths trough FiLM layer: The FiLM layer inside the residual block uses the visual features to modulate the audio feature maps. $\gamma$ and $\beta$ parameters are computed from a dense layer having its input from the visual features.}
    \label{fig:FILM}
\end{figure}

We propose to go further than the "simple" fusion at high level by including a lateral connection to condition audio with vision. Indeed most approaches do not exploit a possible interaction between the different paths. As presented in our previous work \cite{brousmiche2019audio}, the Feature-wise Linear Modulation (FiLM) layer can create a lateral connection between visual and audio paths. We use visual features as input to the FiLM layer to highlight feature maps of the audio modality  (Fig. \ref{fig:FILM}).

More formally, FiLM learns functions $f$ and $h$ to compute $\gamma_{t,c}$ and $\beta_{t,c}$ as a function of input $x_t^1$:

\begin{equation}
    \gamma_{t,c} = f_c(x_t^1) \qquad \beta_{t,c} = h_c(x_t^1)
\end{equation}

$\gamma_{t,c}$ and $\beta_{t,c}$ modulate the activations $\textbf{F}^2_{t,c}$, whose subscripts refer to the $t^{th}$ input and $c^{th}$ audio feature map, via a feature-wise affine transformation:

\begin{equation}
    FiLM(\textbf{F}^2_{t,c}|\gamma_{t,c}, \beta_{t,c}) = \gamma_{t,c} \textbf{F}^2_{t,c} +  \beta_{t,c}
\end{equation}

$f$ and $h$ can be arbitrary functions which are typically implemented with neural networks. FiLM layers allow to manipulate feature maps of a target according to an input by scaling them up or down, negating them, shutting them off, selectively thresholding them (when followed by a ReLU).

\subsection{Audio-visual training}
\label{ssec:training}

Wang \textit{et al.} noticed in \cite{wang2019makes} that multi-modal  networks  are  prone to overfitting due to their increased capacity and different modalities overfit and generalize at different rates. So, they proposed a  complex Gradient-Blending training.  Xiao \textit{et al.} noticed also different dynamic of training depending on the modality \cite{xiao2020audiovisual} and propose to randomly drop the audio path during the training. Unlike \cite{xiao2020audiovisual}, when training unimodal networks, we notice that the audio network need more epoch to reach overfitting compared to the visual network. Therefore, we follow the idea of \cite{xiao2020audiovisual} and randomly drop the upgrade of the visual path, to train more the audio path.
 


\begin{table*}[ht]
    \centering
    \begin{tabular}{c|c|c|c|c|c|c}
        \multicolumn{4}{c}{} & \multicolumn{3}{|c}{Accuracy [\%]}\\
        \cline{5-7}
          & \textbf{model} & \textbf{inputs} & \textbf{pretrained dataset } & \textbf{AVE} & \textbf{UCF51} & \textbf{Kinetics-Sound} \\
        \hline
        \hline
        \multirow{4}{1.5 cm}{End-to-end training} & I3D \cite{carreira2017quo} & V & Kinetics & 73.28 & 86.92 & 80.22\\
        \cline{2-7}
        & R(2+1)D \cite{tran2018closer} & V & Sports1M + Kinetics & 79.19 & 95.54 & 79.10\\
        \cline{2-7}
        & SlowFast \cite{feichtenhofer2019slowfast} & V & Kinetics & 80.41 & 91.78 & 81.91\\
        \cline{2-7}
        & MARS \cite{crasto2019mars} & V & Kinetics & 79.44 & \textbf{97.83} & - \\
        \hline
        \hline
          & \textbf{model} & \textbf{inputs} & \textbf{extraction network} & \textbf{AVE} & \textbf{UCF51} & \textbf{Kinetics-Sound} \\
        \hline
        \hline
        \multirow{4}{1.5 cm}{Feature extraction} & Attention Cluster \cite{long2018attention} & V + A & DenseNet (V) + VGGish (A) & 80.71 & 84.79 & 73.91\\
        \cline{2-7}
         & DMRN \cite{tian2018audio} (our feat.) & V + A & DenseNet (V) + VGGish (A) & 80.96 &  82.93 & 77.5\\
        \cline{2-7}
         & DMRN \cite{tian2018audio} (their feat.) & V + A & VGG19(V)+VGGish-PCA(A) & 85.02 & 81.04 & -\\
        \cline{2-7}
         & MAFnet(our) & V + A & DenseNet (V) + VGGish (A) & \textbf{90.86} & 86.72 & \textbf{83.94}
    \end{tabular}
    \caption{Comparison with state-of-the-art on AVE, UCF51 and Kinetics-Sound datasets. Each model was trained based on code available online. Models are split into two types: end-to-end training and feature extraction. End-to-end training models are trained on larger datasets and then fine-tuned on a smaller dataset. By contrast, feature extraction models are trained on feature previously extracted from the video. Depending on the model, input can be visual frame (V) and/or audio (A). }
    \label{tab:results}
\end{table*}

\section{Experimental results}
\subsection{Datasets}

We evaluate our network on three public datasets: AVE \cite{tian2018audio}, UCF51 \cite{soomro2012ucf101} and Kinetics-Sounds \cite{arandjelovic2017look}.

\textbf{AVE} is a subset of AudioSet \cite{gemmeke2017audio}. The dataset consists of 4143 videos from 28 event classes. Each video lasts 10 s. It covers a wide range of audio-visual events from different domains, e.g., human activities, animal activities, music performances, and vehicle sounds.

\textbf{UCF51} is the second part of the UCF101 dataset  \cite{soomro2012ucf101}. Only the videos of the new 51 classes have sound information. UCF51 dataset consists of 6836 videos from 51 event classes. It concentrates on human actions. The mean video length is 7.0 s. The dataset is partitioned into three splits for training and testing.

\textbf{Kinetics-Sounds} is a subset of the Kinetics dataset \cite{carreira2017quo} and consists of only action classes that are potentially recognizable both visually and aurally. It consists of 21945 videos from 32 events categories. The mean video length is 9.7 s.

\subsection{Feature extraction}

Audio and visual features can easily be extracted from a new video using trained models \cite{tian2018audio, long2018multimodal, lin2019dual, wu2019dual}.  The extracted features are significantly smaller in size than the raw RGB frame and audio data and allow working with smaller networks.

\paragraph{\textbf{Visual feature extraction}} We use an ImageNet pre-trained deep learning model named DenseNet \cite{huang2017densely} to extract visual features from video. The video is split into T clips. As in \cite{tian2018audio}, we choose T = 10, so each clip is one second long without overlapping. For each clip, we extract the output of the DenseNet last convolutional layer for 16 RGB video frames with a global average pooling over the 16 frames to generate one $7\times 7 \times 1920$ dimensional feature map.

\paragraph{\textbf{Audio feature extraction}} We use a VGG-like network \cite{hershey2017cnn} pre-trained on AudioSet to extract audio features. Again, the video is split into T=10 clips of one second each without overlapping. For each clip, we extract the output of the last convolutional layer of the network to generate one $12\times 8 \times 512$ dimensional feature map.

\subsection{Implementation details}

The number of filters in the Residual Block and the number of neurons of the hidden dense layer is 512. Batch normalization is used after hidden dense layers. The network is trained with cross-entropy loss and Adam optimizer with an initial learning rate of 0.001. Early stopping based on the validation accuracy is done, the training is stopped when the validation accuracy does not improve since 50 epochs. During training, we randomly do not update the weights of the visual path. The model is implemented in Tensorflow \cite{abadi2016tensorflow}. 

As UCF51 and Kinetics-Sounds datasets have different video lengths, feature vectors are zero padded to obtain the same length.

\begin{figure*}[ht]
    \centering
    \includegraphics[width=\textwidth]{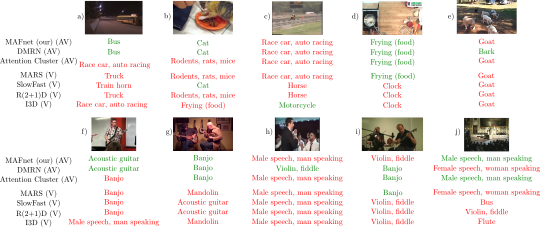}
    \caption{Output prediction of different visual only (V) models and audio-visual (AV) models for some example of the AVE dataset. Each model predicts one class per video. (Green: correct prediction, Red: False prediction)}
    \label{fig:output}
\end{figure*}

\subsection{Event recognition performance}
Table \ref{tab:results} presents event recognition results of MAFnet on AVE, UCF51 and Kinetics-Sounds datasets. We also compare our result with several state-of-the-art methods using different modalities, \textit{i.e.} audio (A) and visual frames (V). For the UCF51 dataset, we report the average accuracy over three testing splits.

MAFnet obtains the best accuracy performance on the AVE dataset among methods based on end-to-end training or feature extraction. End-to-end training methods have the advantage of being trained on larger datasets such as Sports1M or Kinetics to avoid overfitting and then the entire network is fine-tuned on smaller datasets. As the AVE dataset was built as an audio-visual set, the audio information is as important as the visual information. Therefore, as the end-to-end models take into account only the visual information, performances decreases. Furthermore, the AVE dataset includes classes from different events unlike Sports1M and Kinetics datasets which include classes from human activities only. Models based on feature extraction obtain slightly better results than end-to-end training methods due to the use of audio information.

The UCF51 dataset comprises fewer classes with relevant audio information. Indeed, it contains classes that do not produce a particular sound signature or event video with irrelevant background noise. Our network is then not as good as end-to-end training models which take advantage of pretraining on larger datasets and fine-tuning the entire network. On the other hand, models using feature extraction do not fine-tune extractor networks. However, MAFnet is the best model among the architectures that use audio-visual features.

The Kinetics-Sound dataset as weel as the UCF51 dataset is centered on human action but comprises classes potentially recognizable both visually and aurally. Therefore, when the dataset comprises relevant audio and visual information, our network provides the best result. It is capable to take advantage of both modalities. Moreover, it has better integration of the audio and visual information than the other audio-visual models.

Fig. \ref{fig:output} shows examples of output prediction from the different models. For examples a) to e), we observe that the background might impact the choice of the class. The raceway is classified as \textit{Race car, auto racing} even if there is a bus (example a) or a motorcycle (example c). The field with a herd is classified as  \textit{Goat} (example e). Moreover, some specific elements in the video can fool models. The spoon and the plate (example b) may influence the I3D model in the choice of the \textit{Frying (food)} class. In example d), visual models may be fooled by the round shape of the pan. In the case of examples c) and e), the audio modality is not distinctive enough to help the network.

We also note that some instruments can be difficult to distinguish (example f and g) or are occluded (example h). Some videos include several classes but are annotated with only one class (example i). Others are visually indistinctive (example j) but can be classified thanks to the audio modality.

\subsection{Model analysis and discussion}
\label{ssec:analysis}
In this section, we report studies to identify the impact of each module of the MAFnet. We work with the AVE dataset as the dataset assures the presence of the two modalities. We analyze the training method, the impact of the temporal attention, the modality attention and the combination of the two attentions, compare different fusion techniques and the impact of the modality conditioning.

\begin{figure}[t]
    \centering
    \includegraphics[width=0.45\textwidth]{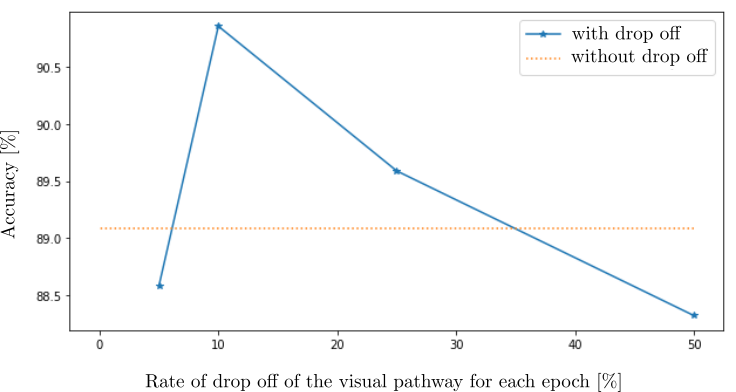}
    \caption{Accuracy of the event recognition of the AVE dataset when using different rates of dropping the weight update of the visual path during training.}
    \label{fig:training}
\end{figure}

\paragraph{\textbf{Training method: Drop off}}
Visual and audio paths do not have the same learning speed. Even without the additional convolutional layers comprised in FiLM, the training of the audio path needs more iterations than for the visual training. Inspired from \cite{xiao2020audiovisual}, we investigate a new multimodal training technique by randomly dropping the update of the visual weights to allow the audio path to train longer. In Fig. \ref{fig:training}, we report the accuracy in function of the dropping rate of the weight update of the visual path. Dropping too often the visual path decreases results compared to training without dropping. We suppose that the visual path is not trained enough. It is also observed that not dropping enough the visual path gives also slightly poorer results. We suppose that the network can not exploit enough sound information. Furthermore, it looses visual information as the visual path is not trained as much as needed.

\paragraph{\textbf{Fusion techniques comparison}}
MAFnet creates a multimodal feature by concatenating the information coming from the visual and audio paths. We analyze the importance of using unimodal information versus multimodal information in the case of event recognition. We also test different fusion techniques present in the literature to determine the best fusion method: addition, concatenation, multimodal compact bilinear pooling (MCB) \cite{gao2016compact} and the multimodal residual fusion (DMR) \cite{tian2018audio}.

As we want to test the fusion techniques, the experiments are made without the FiLM layer. In the case of the unimodal network, the network comprises only the temporal attention module without the modality attention module as only one modality is present.

From the results in Table \ref{tab:fusion}, and as expected, we conclude that the dataset is easier to classify using visual information only than sound information only. Multimodal information increases the performance compared to unimodal. Concatenation has the best result and is even slightly better than more complex fusion techniques like MCB or DMR.

\begin{table}[t]
    \centering
    \begin{tabular}{c|c}
        fusion technique & Accuracy [\%] \\
        \hline
        \hline
        visual & 75.63  \\
        \hline
        audio & 69.29  \\
        \hline
        \hline
        addition &  84.77 \\
        \hline
        concatenation &  \textbf{89.34} \\
        \hline
        MCB & 88.83 \\
        \hline
        DMR & 87.56 \\
    \end{tabular}
    \caption{Comparison unimodal versus multimodal event recognition and the use of different fusion techniques on AVE dataset.}
    \label{tab:fusion}
\end{table}


\paragraph{\textbf{Attention analysis}}
In this section, we analyze the impact of each attention module. Table \ref{tab:attention} presents the event recognition results without attention, with temporal attention only and with modality attention only. Again, the network does not comprise the FiLM layer for this ablation study. The temporal attention allows to take into account the temporal context and dynamically highlights particular time windows. Not each time window comprises relevant information for the classification. The modality attention highlights a modality. Indeed, depending on the video a modality can have more importance than the other. Each attention module has a positive impact on the accuracy and the combination of both attentions gets the best result. 

\begin{table}[t]
    \centering
    \begin{tabular}{c|c}
        Attention type & Accuracy [\%] \\
        \hline
        \hline
        without attention & 87.82 \\
        \hline
        temporal attention & 88.92 \\
        \hline
        modality attention & 88.66 \\
        \hline
        modality \& temporal attention & \textbf{89.34} \\
    \end{tabular}
    \caption{Ablation study of the modality \& temporal attention module on the AVE dataset.}
    \label{tab:attention}
\end{table}

\paragraph{\textbf{Modality conditioning analysis}}

\begin{table}[t]
    \centering
    \begin{tabular}{c|c}
        FiLM location & Accuracy [\%] \\
        \hline
        \hline
        Add residual block without & \\
        FILM in both path & 86.55 \\
        \hline
        FiLM layer in both paths & 87.62 \\
        \hline
        FiLM layer in audio path & \textbf{90.86} \\
        \hline
        FiLM layer in visual path & 90.61 \\
        
    \end{tabular}
    \caption{Evaluation of the lateral connection between visual and audio paths with FiLM layer on the AVE dataset.}
    \label{tab:film}
\end{table}

\begin{figure}[t]
    \centering
    \includegraphics[width=0.5\textwidth]{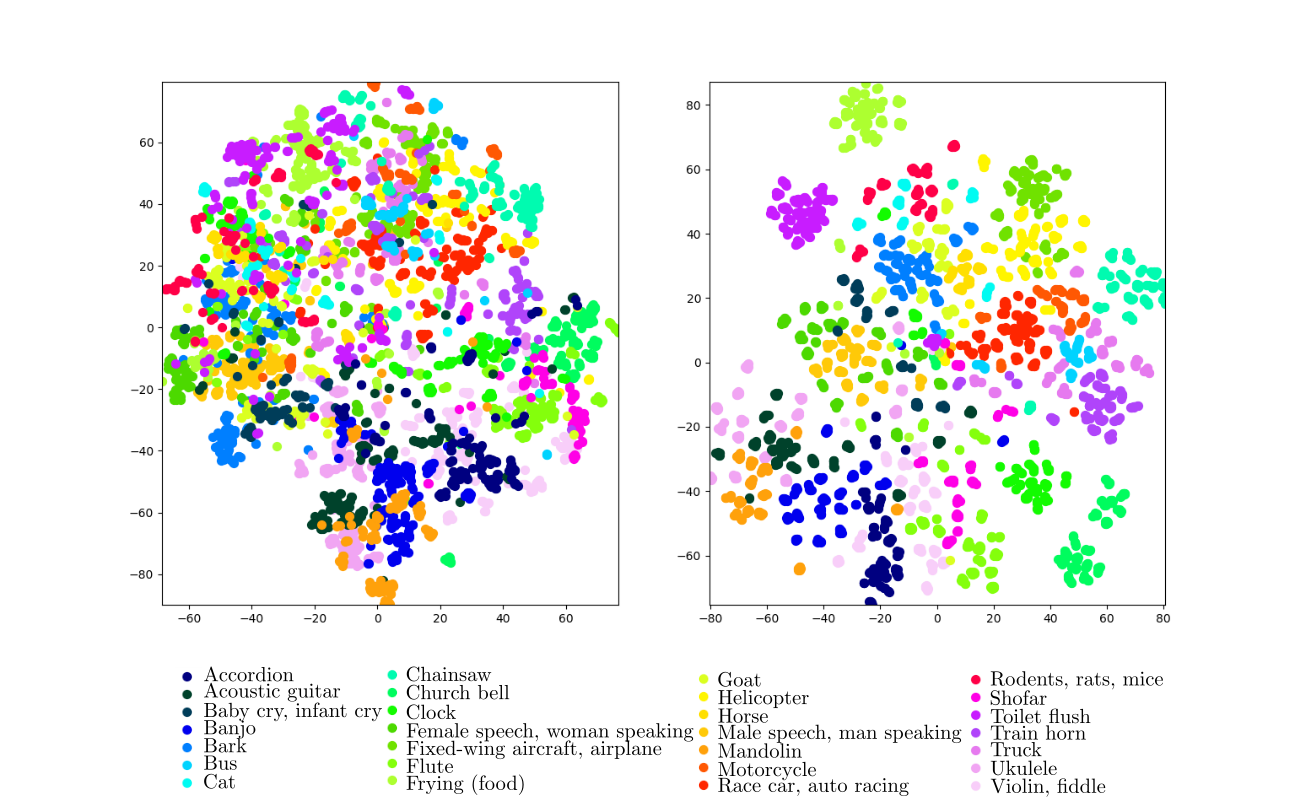}
    \caption{T-SNE visualization of the embedding of the residual block just before (left) and after (right) the FiLM layer in the audio path.}
    \label{fig:embedding}
\end{figure}

In this section, we analyze the impact of the lateral connection, the modality conditioning (Table \ref{tab:film}). It is observed that adding FiLM in visual and audio path provides better results than without any conditioning. However, conditioning only one modality is better than conditioning both modalities whatever the conditioned modality. The best result is obtained when visual conditions audio. We suppose that vision contributes more to audio than the other way. This is in accordance with previous observations in \cite{brousmiche2019audio}.


Fig \ref{fig:embedding} compares the embedding of the residual block just before and after the FiLM layer in the audio path (Fig. \ref{fig:FILM}). We use average pooling and t-SNE \cite{maaten2008visualizing} to reduce the embedding dimension to 2D. We observe a better clustering of the different classes after including the visual information in the audio path.



\section{Conclusion}
We proposed the Multi-level Attention Fusion (MAFnet) in the context of event recognition task. Our network includes a modality \& temporal attention module. It dynamically associates a score to each modality at each time window to highlight the relevant modality and time window. To go further than a simple late fusion, we condition one modality with the other with a FiLM layer. It highlights selected audio feature maps based on visual information. Finally, to take into account the different learning dynamics of each modality, we randomly drop the weight update of the visual path. We evaluate our network on three datasets and achieve better accuracy than the current audio-visual models. 

MAFnet shows promising results for audiovisual data fusion in the context of event classification. When exploiting audio-visual data, the fusion of the two modalities is not the only important element, the conditioning between the modalities paths is necessary to make the best use of the audio-visual information. Conditioning with the FiLM layer modifies the hidden representation of audio modalities based on visual information. In view of the conditioning results, future research should investigate the interaction between the audio and visual path at different levels of the architecture as well as study different conditioning methods.

\section*{Acknowlgedgments}
This work was funded by CHISTERA IGLU project and the European Regional Development Fund (ERDF). The author would also like to thank the NVIDIA Corporation for donating GPUs. 
\bibliography{mybibfile}

\begin{thebibliography}{10}
\expandafter\ifx\csname url\endcsname\relax
  \def\url#1{\texttt{#1}}\fi
\expandafter\ifx\csname urlprefix\endcsname\relax\def\urlprefix{URL }\fi
\expandafter\ifx\csname href\endcsname\relax
  \def\href#1#2{#2} \def\path#1{#1}\fi

\bibitem{dufour2012intelligent}
J.-Y. Dufour, Intelligent video surveillance systems, John Wiley \& Sons, 2012.

\bibitem{kala2016road}
R.~Kala, {On-road intelligent vehicles: Motion planning for intelligent
  transportation systems}, Butterworth-Heinemann, 2016.

\bibitem{d2010review}
T.~D’Orazio, M.~Leo, A review of vision-based systems for soccer video
  analysis, Pattern Recognition 43~(8) (2010) 2911--2926.

\bibitem{gibbon2008introduction}
D.~C. Gibbon, Z.~Liu, Introduction to video search engines, Springer Science \&
  Business Media, 2008.

\bibitem{real2017youtube}
E.~Real, J.~Shlens, S.~Mazzocchi, X.~Pan, V.~Vanhoucke, {Youtube-boundingboxes:
  A large high-precision human-annotated data set for object detection in
  video}, in: Proceedings of the IEEE Conference on Computer Vision and Pattern
  Recognition (CVPR), 2017, pp. 5296--5305.

\bibitem{carreira2017quo}
J.~Carreira, A.~Zisserman, Quo vadis, action recognition? a new model and the
  kinetics dataset, in: Proceedings of the IEEE Conference on Computer Vision
  and Pattern Recognition (CVPR), 2017, pp. 6299--6308.

\bibitem{karpathy2014large}
A.~Karpathy, G.~Toderici, S.~Shetty, T.~Leung, R.~Sukthankar, L.~Fei-Fei,
  Large-scale video classification with convolutional neural networks, in:
  Proceedings of the IEEE Conference on Computer Vision and Pattern Recognition
  (CVPR), 2014, pp. 1725--1732.

\bibitem{simonyan2014very}
K.~Simonyan, A.~Zisserman, Very deep convolutional networks for large-scale
  image recognition, arXiv preprint arXiv:1409.1556.

\bibitem{szegedy2016rethinking}
C.~Szegedy, V.~Vanhoucke, S.~Ioffe, J.~Shlens, Z.~Wojna, Rethinking the
  inception architecture for computer vision, in: Proceedings of the IEEE
  Conference on Computer Vision and Pattern Recognition (CVPR), 2016, pp.
  2818--2826.

\bibitem{simonyan2014two}
K.~Simonyan, A.~Zisserman, Two-stream convolutional networks for action
  recognition in videos, in: Advances in Neural Information Processing Systems
  (NIPS), 2014, pp. 568--576.

\bibitem{donahue2015long}
J.~Donahue, L.~Anne~Hendricks, S.~Guadarrama, M.~Rohrbach, S.~Venugopalan,
  K.~Saenko, T.~Darrell, Long-term recurrent convolutional networks for visual
  recognition and description, in: Proceedings of the IEEE Conference on
  Computer Vision and Pattern Recognition (CVPR), 2015, pp. 2625--2634.

\bibitem{tran2015learning}
D.~Tran, L.~Bourdev, R.~Fergus, L.~Torresani, M.~Paluri, {Learning
  spatiotemporal features with 3D convolutional networks}, in: Proceedings of
  the IEEE International Conference on Computer Vision (ICCV), 2015, pp.
  4489--4497.

\bibitem{kazakos2019epic}
E.~Kazakos, A.~Nagrani, A.~Zisserman, D.~Damen, {Epic-fusion: Audio-visual
  temporal binding for egocentric action recognition}, in: Proceedings of the
  IEEE International Conference on Computer Vision (ICCV), 2019, pp.
  5492--5501.

\bibitem{arandjelovic2017look}
R.~Arandjelovic, A.~Zisserman, Look, listen and learn, in: Proceedings of the
  IEEE International Conference on Computer Vision (ICCV), 2017, pp. 609--617.

\bibitem{arandjelovic2018objects}
R.~Arandjelovic, A.~Zisserman, Objects that sound, in: Proceedings of the
  European Conference on Computer Vision (ECCV), 2018, pp. 435--451.

\bibitem{aytar2016soundnet}
Y.~Aytar, C.~Vondrick, A.~Torralba, {Soundnet: Learning sound representations
  from unlabeled video}, in: Advances in Neural Information Processing Systems
  (NIPS), 2016, pp. 892--900.

\bibitem{owens2018audio}
A.~Owens, A.~A. Efros, Audio-visual scene analysis with self-supervised
  multisensory features, in: Proceedings of the European Conference on Computer
  Vision (ECCV), 2018, pp. 631--648.

\bibitem{owens2016ambient}
A.~Owens, J.~Wu, J.~H. McDermott, W.~T. Freeman, A.~Torralba, Ambient sound
  provides supervision for visual learning, in: Proceedings of the European
  Conference on Computer Vision (ECCV), Springer, 2016, pp. 801--816.

\bibitem{tian2018audio}
Y.~Tian, J.~Shi, B.~Li, Z.~Duan, C.~Xu, Audio-visual event localization in
  unconstrained videos, in: Proceedings of the European Conference on Computer
  Vision (ECCV), 2018, pp. 247--263.

\bibitem{tian2019multi}
Y.~Tian, Y.~Cao, J.~Wu, W.~Hu, C.~Song, T.~Yang, Multi-cue combination network
  for action-based video classification, IET Computer Vision 13~(6) (2019)
  542--548.

\bibitem{long2018multimodal}
X.~Long, C.~Gan, G.~De~Melo, X.~Liu, Y.~Li, F.~Li, S.~Wen, Multimodal keyless
  attention fusion for video classification, in: Thirty-Second AAAI Conference
  on Artificial Intelligence, 2018.

\bibitem{soomro2012ucf101}
K.~Soomro, A.~R. Zamir, M.~Shah, {UCF101: A dataset of 101 human actions
  classes from videos in the wild}, arXiv preprint arXiv:1212.0402.

\bibitem{yue2015beyond}
J.~Yue-Hei~Ng, M.~Hausknecht, S.~Vijayanarasimhan, O.~Vinyals, R.~Monga,
  G.~Toderici, {Beyond short snippets: Deep networks for video classification},
  in: Proceedings of the IEEE Conference on Computer Vision and Pattern
  Recognition (CVPR), 2015, pp. 4694--4702.

\bibitem{ma2019ts}
C.-Y. Ma, M.-H. Chen, Z.~Kira, G.~AlRegib, {TS-LSTM and temporal-inception:
  Exploiting spatiotemporal dynamics for activity recognition}, Signal
  Processing: Image Communication 71 (2019) 76--87.

\bibitem{li2018videolstm}
Z.~Li, K.~Gavrilyuk, E.~Gavves, M.~Jain, C.~G. Snoek, {VideoLSTM convolves,
  attends and flows for action recognition}, Computer Vision and Image
  Understanding 166 (2018) 41--50.

\bibitem{tran2018closer}
D.~Tran, H.~Wang, L.~Torresani, J.~Ray, Y.~LeCun, M.~Paluri, A closer look at
  spatiotemporal convolutions for action recognition, in: Proceedings of the
  IEEE Conference on Computer Vision and Pattern Recognition (CVPR), 2018, pp.
  6450--6459.

\bibitem{xie2018rethinking}
S.~Xie, C.~Sun, J.~Huang, Z.~Tu, K.~Murphy, {Rethinking spatiotemporal feature
  learning: Speed-accuracy trade-offs in video classification}, in: Proceedings
  of the European Conference on Computer Vision (ECCV), 2018, pp. 305--321.

\bibitem{feichtenhofer2016convolutional}
C.~Feichtenhofer, A.~Pinz, A.~Zisserman, Convolutional two-stream network
  fusion for video action recognition, in: Proceedings of the IEEE Conference
  on Computer Vision and Pattern Recognition (CVPR), 2016, pp. 1933--1941.

\bibitem{wang2016temporal}
L.~Wang, Y.~Xiong, Z.~Wang, Y.~Qiao, D.~Lin, X.~Tang, L.~Van~Gool, {Temporal
  segment networks: Towards good practices for deep action recognition}, in:
  European Conference on Computer Vision, Springer, 2016, pp. 20--36.

\bibitem{long2018attention}
X.~Long, C.~Gan, G.~De~Melo, J.~Wu, X.~Liu, S.~Wen, {Attention clusters: Purely
  attention based local feature integration for video classification}, in:
  Proceedings of the IEEE Conference on Computer Vision and Pattern Recognition
  (CVPR), 2018, pp. 7834--7843.

\bibitem{gao2016compact}
Y.~Gao, O.~Beijbom, N.~Zhang, T.~Darrell, Compact bilinear pooling, in:
  Proceedings of the IEEE Conference on Computer Vision and Pattern
  Recognition, 2016, pp. 317--326.

\bibitem{lin2019dual}
Y.-B. Lin, Y.-J. Li, Y.-C.~F. Wang, Dual-modality seq2seq network for
  audio-visual event localization, in: IEEE International Conference on
  Acoustics, Speech and Signal Processing (ICASSP), IEEE, 2019, pp. 2002--2006.

\bibitem{wu2019dual}
Y.~Wu, L.~Zhu, Y.~Yan, Y.~Yang, Dual attention matching for audio-visual event
  localization, in: Proceedings of the IEEE International Conference on
  Computer Vision (ICCV), 2019, pp. 6292--6300.

\bibitem{xiao2020audiovisual}
F.~Xiao, Y.~J. Lee, K.~Grauman, J.~Malik, C.~Feichtenhofer, {Audiovisual
  SlowFast networks for video recognition}, arXiv preprint arXiv:2001.08740.

\bibitem{mnih2014recurrent}
V.~Mnih, N.~Heess, A.~Graves, et~al., Recurrent models of visual attention, in:
  Advances in Neural Information Processing Systems (NIPS), 2014, pp.
  2204--2212.

\bibitem{vaswani2017attention}
A.~Vaswani, N.~Shazeer, N.~Parmar, J.~Uszkoreit, L.~Jones, A.~N. Gomez,
  {\L}.~Kaiser, I.~Polosukhin, Attention is all you need, in: Advances in
  Neural Information Processing Systems (NIPS), 2017, pp. 5998--6008.

\bibitem{wang2020cascade}
J.~Wang, X.~Peng, Y.~Qiao, Cascade multi-head attention networks for action
  recognition, Computer Vision and Image Understanding (2020) 102898.

\bibitem{li2020spatio}
J.~Li, X.~Liu, W.~Zhang, M.~Zhang, J.~Song, N.~Sebe, Spatio-temporal attention
  networks for action recognition and detection, IEEE Transactions on
  Multimedia.

\bibitem{huang2017arbitrary}
X.~Huang, S.~Belongie, Arbitrary style transfer in real-time with adaptive
  instance normalization, in: Proceedings of the IEEE International Conference
  on Computer Vision (ICCV), 2017, pp. 1501--1510.

\bibitem{kim2017dynamic}
T.~Kim, I.~Song, Y.~Bengio, Dynamic layer normalization for adaptive neural
  acoustic modeling in speech recognition, in: Proceedings of Interspeech,
  2017, pp. 2655--2659.

\bibitem{de2017modulating}
H.~De~Vries, F.~Strub, J.~Mary, H.~Larochelle, O.~Pietquin, A.~C. Courville,
  Modulating early visual processing by language, in: Advances in Neural
  Information Processing Systems (NIPS), 2017, pp. 6594--6604.

\bibitem{abdelnour2019visual}
J.~Abdelnour, G.~Salvi, J.~Rouat, From visual to acoustic question answering,
  arXiv preprint arXiv:1902.11280.

\bibitem{brousmiche2019audio}
M.~Brousmiche, J.~Rouat, S.~Dupont, Audio-visual fusion and conditioning with
  neural networks for event recognition, in: IEEE International Workshop on
  Machine Learning for Signal Processing (MLSP), IEEE, 2019, pp. 1--6.

\bibitem{perez2018film}
E.~Perez, F.~Strub, H.~De~Vries, V.~Dumoulin, A.~Courville, Film: Visual
  reasoning with a general conditioning layer, in: Thirty-Second AAAI
  Conference on Artificial Intelligence, 2018.

\bibitem{huang2017densely}
G.~Huang, Z.~Liu, L.~Van Der~Maaten, K.~Q. Weinberger, Densely connected
  convolutional networks, in: Proceedings of the IEEE Conference on Computer
  Vision and Pattern Recognition, 2017, pp. 4700--4708.

\bibitem{hershey2017cnn}
S.~Hershey, S.~Chaudhuri, D.~P. Ellis, J.~F. Gemmeke, A.~Jansen, R.~C. Moore,
  M.~Plakal, D.~Platt, R.~A. Saurous, B.~Seybold, et~al., {CNN architectures
  for large-scale audio classification}, in: IEEE International Conference on
  Acoustics, Speech and Signal Processing (ICASSP), IEEE, 2017, pp. 131--135.

\bibitem{goldstein2016sensation}
E.~B. Goldstein, J.~Brockmole, Sensation and perception, Cengage Learning,
  2016.

\bibitem{zhou2019modality}
P.~Zhou, W.~Yang, W.~Chen, Y.~Wang, J.~Jia, Modality attention for end-to-end
  audio-visual speech recognition, in: IEEE International Conference on
  Acoustics, Speech and Signal Processing (ICASSP), IEEE, 2019, pp. 6565--6569.

\bibitem{wang2019makes}
W.~Wang, D.~Tran, M.~Feiszli, What makes training multi-modal networks hard?,
  arXiv preprint arXiv:1905.12681.

\bibitem{feichtenhofer2019slowfast}
C.~Feichtenhofer, H.~Fan, J.~Malik, K.~He, {SlowFast networks for video
  recognition}, in: Proceedings of the IEEE International Conference on
  Computer Vision (ICCV), 2019, pp. 6202--6211.

\bibitem{crasto2019mars}
N.~Crasto, P.~Weinzaepfel, K.~Alahari, C.~Schmid, {MARS: Motion-augmented RGB
  stream for action recognition}, in: Proceedings of the IEEE Conference on
  Computer Vision and Pattern Recognition (CVPR), 2019, pp. 7882--7891.

\bibitem{gemmeke2017audio}
J.~F. Gemmeke, D.~P. Ellis, D.~Freedman, A.~Jansen, W.~Lawrence, R.~C. Moore,
  M.~Plakal, M.~Ritter, {Audio set: An ontology and human-labeled dataset for
  audio events}, in: IEEE International Conference on Acoustics, Speech and
  Signal Processing (ICASSP), IEEE, 2017, pp. 776--780.

\bibitem{abadi2016tensorflow}
M.~Abadi, P.~Barham, J.~Chen, Z.~Chen, A.~Davis, J.~Dean, M.~Devin,
  S.~Ghemawat, G.~Irving, M.~Isard, et~al., {Tensorflow: A system for
  large-scale machine learning}, in: 12th $\{$USENIX$\}$ Symposium on Operating
  Systems Design and Implementation ($\{$OSDI$\}$ 16), 2016, pp. 265--283.

\bibitem{maaten2008visualizing}
L.~v.~d. Maaten, G.~Hinton, {Visualizing data using t-SNE}, Journal of machine
  learning research 9~(Nov) (2008) 2579--2605.

\end{thebibliography}

\end{document}